\documentclass[conference]{IEEEtran}
\IEEEoverridecommandlockouts
\usepackage{cite}
\usepackage{amsmath,amssymb,amsfonts}
\usepackage{algorithmic}
\usepackage{graphicx}
\usepackage{textcomp}
\usepackage{xcolor}
\usepackage{algorithm}
\usepackage{multirow}
\usepackage{subcaption}
\usepackage{colortbl}

\def\BibTeX{{\rm B\kern-.05em{\sc i\kern-.025em b}\kern-.08em
    T\kern-.1667em\lower.7ex\hbox{E}\kern-.125emX}}
\begin{document}

\title{A Multi-constraint and Multi-objective Allocation Model for Emergency Rescue in IoT Environment}

\author{
\IEEEauthorblockN{Xinrun Xu\textsuperscript{1,2}$^\text{*}$, Zhanbiao Lian\textsuperscript{1,2}$^\text{*}$, Yurong Wu\textsuperscript{1,2}, Manying Lv\textsuperscript{1,2}, Zhiming Ding\textsuperscript{1,2}, Jin Yan\textsuperscript{1,2}$^\text{\dag}$, Shan Jiang\textsuperscript{3}$^\text{\dag}$}
\IEEEauthorblockA{
\textsuperscript{1}{University of Chinese Academy of Sciences}, Beijing, China,\\
\textsuperscript{2}{Institute of Software, Chinese Academy of Sciences}, Beijing, China \\
\textsuperscript{3}{Advanced Institute of Big Data (Beijing), Beijing, China}
}
Email:\{xuxinrun20, lianzhanbiao21, yvette.yan\}@mails.ucas.ac.cn, jiangshan@alumni.nudt.edu.cn
}
\maketitle
\renewcommand{\thefootnote}{\fnsymbol{footnote}}
\footnotetext[1]{Xinrun Xu and Zhanbiao Lian have contributed equally to this work.}
\footnotetext[2]{Jin Yan and Shan Jiang are Corresponding authors.}
\renewcommand{\thefootnote}{\arabic{footnote}}

\begin{abstract}
Emergency relief operations are essential in disaster aftermaths, necessitating effective resource allocation to minimize negative impacts and maximize benefits. In prolonged crises or extensive disasters, a systematic, multi-cycle approach is key for timely and informed decision-making. Leveraging advancements in IoT and spatio-temporal data analytics, we've developed the Multi-Objective Shuffled Gray-Wolf Frog Leaping Model (MSGW-FLM). This multi-constraint, multi-objective resource allocation model has been rigorously tested against 28 diverse challenges, showing superior performance in comparison to established models such as NSGA-II, IBEA, and MOEA/D. MSGW-FLM's effectiveness is particularly notable in complex, multi-cycle emergency rescue scenarios, which involve numerous constraints and objectives. This model represents a significant step forward in optimizing resource distribution in emergency response situations.
\end{abstract}


\section{Introduction}
In light of recent advancements in information technology\cite{powers2023using}, particularly the surge in artificial intelligence\cite{07KamalPaul2022}, IoT sensing\cite{santhanaraj2023internet}, and the evolution of smart cities\cite{pandiyan2023technological}, Spatio-temporal Artificial Intelligence offers fresh insights for addressing emergency rescue challenges\cite{khan2023systematic}. The integration of these technologies into post-disaster emergency response has garnered significant attention in recent years\cite{aringhieri2022fairness, munawar2022disruptive}.
Emergency response in the aftermath of a disaster is critical to reducing casualties and protecting lives and assets. The challenge is to analyse complex information to develop an effective rescue strategy. Improving response effectiveness requires data on individual behaviour, aggregation and spatio-temporal patterns in disaster zones. Spatio-temporal IoT data provides a comprehensive view of human, machine and object interactions, providing decision-makers with rich semantic information for informed decision-making.

This study delves into the pivotal role of IoT-driven spatio-temporal data in underpinning effective decision-making for disaster relief, particularly in the nuanced arena of emergency resource allocation.
Emergency managers harness the power of IoT to gather real-time data \cite{huang2023machine} about pedestrian distribution at the disaster site and its vicinity, allowing for an accurate initial estimation of required emergency resources, including rescue supplies, vehicles, and personnel.
As the situation evolves, IoT devices continuously capture critical post-disaster data \cite{zeng2023sensors}. When updated information about the catastrophe's progression is relayed, emergency decision-makers can refine and recalibrate their rescue plans, enhancing the efficacy of subsequent rescue cycles.
This leads to a dynamic response scenario. Emergency response, enhanced by IoT devices, involves dynamic, multi-cycle assessments and decisions. As rescue efforts progress, real-time disaster data from IoT devices becomes increasingly valuable, ensuring that subsequent rescue efforts are better informed and more effective. In the IoT age, emergency response is a data-driven, iterative process.
Our main contributions are outlined as follows:
\begin{enumerate}
    \item In the context of multi-cycle emergency rescue within IoT framework, we introduce the MSGW-FLM: a multi-constraint multi-objective emergency resource allocation model derived from a combination of the grey wolf optimization and frog leaping algorithms. To our knowledge, this pioneering effort represents the inaugural exploration of dynamic multi-cycle emergency rescue scenarios. 
    \item By harnessing spatio-temporal data, we conducted rigorous experiments on 28 multi-objective subtasks using four evaluation metrics: HV (hyper volume)\cite{153zitzler1999multiobjective}, IGD (inverted generation distance)\cite{154durillo2011jmetal}, Spread, and convergence, benchmarking against established models NSGA-II, IBEA, and MOEA/D. Our findings consistently demonstrate the superiority of our approach over conventional models in the majority of evaluations.
\end{enumerate}

\section{MSGW-FLM: multi-objective shuffled grey wolf-frog leaping model}

\begin{figure}[!t]
\centering
\includegraphics[width=\linewidth]{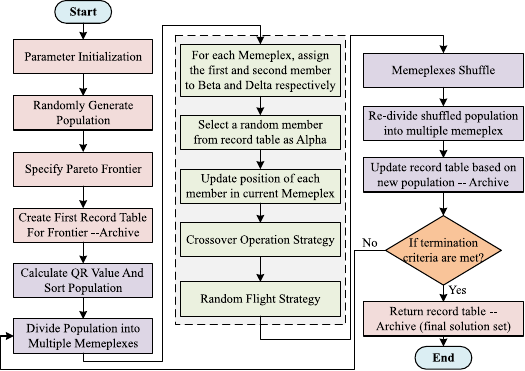}
\caption{Framework of MSGW-FLM}
\label{fig:04}
\end{figure}

This paper introduces a solution for the Constraint Multi-objective Decision-making Optimization Problem (CMDOP) \cite{feng2022hybrid}. We present a strategy combining the Grey Wolf Optimization Algorithm (GWOA) \cite{mirjalili2014grey,zamfirache2022policy,liu2023agricultural} and Shuffled Frog Leaping Algorithm (SFLA) \cite{eusuff2003optimization,yang2022stochastic,chen2022multi}, termed the Multi-objective Shuffled Grey Wolf-Frog Leaping Model (MSGW-FLM), displayed in Fig. \ref{fig:04}.

\subsection{Shuffled Frog Leaping Algorithm (SFLA)}

The Shuffled Frog Leaping Algorithm (SFLA) \cite{eusuff2003optimization}, is a meta-heuristic approach inspired by the foraging behavior of frogs. In SFLA, frogs, representing potential solutions, are grouped into memeplexes. Each memeplex is a sub-group containing multiple frogs, which influence each other to improve their positions. Within a memeplex, the position of the least optimal frog is adjusted towards that of the most optimal frog, simulating information exchange and solution enhancement. Frogs are reallocated across different memeplexes in each generation, encouraging the exploration of new solutions. If a frog fails to find a better position, it executes a random leap to avoid local optima. In each generation, frogs may jump between 1 to 3 times, but the total number of jumps can be less than the number of frogs if the available 'rocks' (representing positions in the solution space) are limited. The main process of SFLA is outlined in Algorithm \ref{algo:01}.

The Memetic Evolutionary (ME) formula is:

\begin{equation}
    \label{eq10}
    Y_k=\left\{\left(X_i\right)_k \mid X_i=X(k+m *(i-1)) \right\}
\end{equation}
where $i=1,2, \cdots, n.\ k=1,2, \cdots, m$, $m$ denotes memeplex count and $n$ is members per memeplex. Use ME to balance member allocation. Afterwards, memeplexes are combined and shuffled. Iteration continues until termination.

\begin{algorithm}
\caption{Shuffled Frog Leaping Algorithm}
\label{algo:01}
\begin{algorithmic}[1]
    \STATE Parameter initialization;
    \STATE Randomly generate populations $X$;
    \STATE Calculate the target fitness function value $F$;
    \REPEAT
    \STATE $F_x \leftarrow Sort(F,X)$; // Sort the population in ascending order according to the fitness function value 
    \STATE $Y_{memeplex} \leftarrow Seprate (F_x)$; // Divide the sorted population into memeplex
    \FOR{$me \in Y_{memeplex} $}
    \STATE $Y_{memeplex}\cdot me \leftarrow MemeticEvolutionary(me)$; // Memetic evolution for each memeplex
    \ENDFOR
    \STATE $Y_{memeplex}\leftarrow Shuffle(Y_{memeplex})$;
    \IF{($getMax\ ||\ Flag(termination\ criteria)$)}
    \RETURN $global best$;
    \ENDIF
    \UNTIL reach MAX; // Maximum fitness evaluation size 
\end{algorithmic}
\end{algorithm}

\begin{algorithm}[!h]
	\caption{Grey Wolf Optimization Algorithm}
	\begin{algorithmic}[1]
	    \STATE Parameter initialization;
	    \STATE Randomly generate populations $X$;
	    \STATE Calculate the target fitness function value $F$;
	    \STATE Assign Alpha-$\alpha$, Beta-$\beta$, and Delta-$\delta$ role to grey wolves;\\
	    \REPEAT
	    \STATE $L_X\leftarrow Update(F,X)$; // Update locations\\
	    \STATE $F\leftarrow Fitness(L_X,X)$; // Calculate the fitness function value corresponding to the new position \\
	    \STATE $(X \cdot \alpha, X \cdot \beta, X \cdot \delta) \leftarrow$ Update $(\alpha, \beta, \delta)$; // Update roles\\
	    \IF{($getMax\ ||\ Flag(termination\ criteria)$)}
	    \RETURN $\alpha$;
	    \ENDIF
	    \UNTIL{reach MAX;}\ // Maximum fitness evaluation size)
	\end{algorithmic}
	\label{algo:02}
\end{algorithm}

\subsection{Grey Wolf Optimization Algorithm (GWOA)}

The Grey Wolf Optimization Algorithm (GWOA) \cite{mirjalili2014grey} is inspired by the rank and hunting scenarios of grey wolves. It's an iterative, population-based meta-heuristic optimization with strong convergence, minimal parameters, and straightforward implementation. 
GWOA uses wolves' strict social hierarchy and hunting strategies for its mathematical model. The overall flow is in Algorithm \ref{algo:02}, comprising four phases: social hierarchy, encircling, hunting, and attacking prey.

\textbf{Social Hierarchy: }
In order of hierarchy, grey wolves have roles: Alpha-$\alpha$, Beta-$\beta$, Delta-$\delta$, and Omega-$\omega$. The Alpha leads in decisions like hunting and resting, while Beta and Delta oversee Omega wolves and assist Alpha in decision-making.

\textbf{Encircling prey: }
Grey wolves encircle their prey during the search, which can be mathematically modelled as:

\begin{align}
    \label{eq12}
    \vec{X}^{\prime} &= \vec{X}_p - \vec{A} \cdot \vec{D} \\
    \label{eq13}
    \vec{D} &= \left|\vec{C} \cdot \vec{X}_p - \vec{X}\right|\\
    \label{eq14}
    \vec{A} &=2 a \cdot \vec{r}_1-a\\
    \label{eq15}
    \vec{C} &=2 \vec{r}_2
\end{align}
where $\vec{A}$, $\vec{C}$ are coefficient vectors; $\vec{X}_p$ is the prey's position; $\vec{X}$ is the grey wolves' current position; $\vec{D}$ is the distance vector. $a$ linearly decreases from 2 to 0 per iteration, and $\vec{r_1}$, $\vec{r_2}$ are random vectors in [0, 1].

\begin{figure}[!t]
\centering
\includegraphics[width=\linewidth]{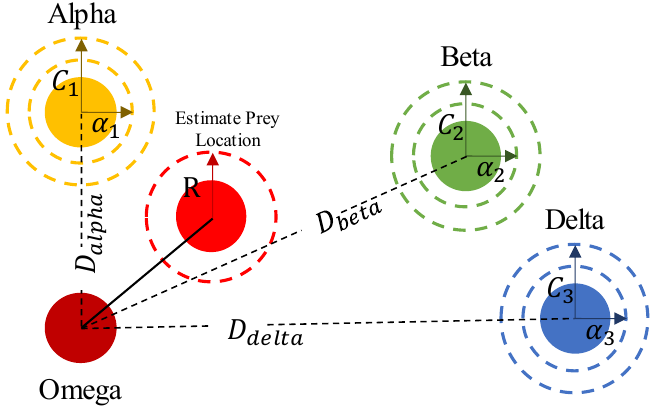}
\caption{Diagram of grey wolves' location update.}
\label{fig:03}
\end{figure}

\textbf{Hunting prey:}
Grey wolves identify potential prey (optimal solution) locations, with Alpha, Beta, and Delta wolves guiding the search. Omega wolves update their positions based on these leading wolves, moving towards a random position within their circle. They advance toward potential prey. Fig. \ref{fig:03} depicts the position updates of Alpha, Beta, and Delta wolves, which guide other wolves in random position updates. Wolf-hunting behaviour is modelled as:

\begin{align}
    \label{eq16}
    \vec{D}_a &=\left|\vec{C}_1 \cdot \vec{X}_a-\vec{X}\right|\\
    \label{eq17}
    \vec{D}_\beta &=\left|\vec{C}_2 \cdot \vec{X}_\beta-\vec{X}\right|\\
    \label{eq18}
    \vec{D}_\delta &=\left|\vec{C}_3 \cdot \vec{X}_{\bar{\delta}}-\vec{X}\right|\\
     \label{eq19}
    \vec{X}_1 &=\vec{X}_a-\vec{A}_1 \cdot \vec{D}_a\\
    \label{eq20}
    \vec{X}_2 &=\vec{X}_\beta-\vec{A}_2 \cdot \vec{D}_a\\
     \label{eq21}
    \vec{X}_3 &=\vec{X}_\delta-\vec{A}_3 \cdot \vec{D}_a\\
    \label{eq22}
    \vec{X}^{\prime} &=\frac{1}{3}\left(\vec{X}_1+\vec{X}_2+\vec{X}_3\right)
\end{align}

\textbf{Attacking prey:}
During the hunting finale, grey wolves decide to attack or seek better prey. The reduction in vector $\vec{A}$ depends on $a$. As $a$ decreases, $\vec{A}$ randomly falls between $[-2a,2a]$, making wolves approach the leader. If $|\vec{A}|<1$, they attack prey (local optimum). If $|\vec{A}|>1$, they seek better prey (global optimum).

\subsection{Multi-objective Shuffled Grey Wolf-Frog Leaping Model (MSGW-FLM)}
This study integrates SFLA's memeplex structure into GWOA for efficient solution space exploration. 
To enhance ranking precision in multi-objective problems, we utilize NSGA-II's \cite{babalik2018multi} fast non-dominated sorting and crowding distance to determine a Quality Rank (QR) for population sorting. Utilize integration of the crossover operator \cite{anderson1999genetic} accelerates convergence and optimizes efficiency, while using the Levy flight strategy \cite{yang2009cuckoo} dynamically determines the step size, fostering adaptive solutions.
The MSGW-FLM flow chart and pseudocode are in Fig. \ref{fig:04} and Algorithm \ref{algo:03}, respectively.

\begin{algorithm}[!h]
	\caption{The Pseudocode of Multi-objective Shuffled Grey Wolf-Frog Leaping Model}
	\begin{algorithmic}[1]
	    \STATE Set the parameters of the algorithm;
	    \STATE Generate a random population within boundaries;
	    \STATE Calculate the value of the objective functions for each member in the population;
		\STATE Separate the population into frontiers;
		\STATE Create the first archive from $l^{st}$ frontier;
        \STATE \textbf{for} $t = P$ to $MAXFes$: // $P$ is the population size
		\STATE \textbf{Repeat }(until reach $MAXFes$) // $MAXFes$: Maximum fitness evaluation size
		\STATE \ \ \ \ Calculate $QR$ value of the members;
		\STATE \ \ \ \ Separate the population into memeplexes;
		\STATE \ \ \ \ \textbf{for} $i = l$ to $m$: // $m$ memeplex number
		\STATE \ \ \ \ \textbf{Repeat} (for each memeplex)
		\STATE \ \ \ \ \ \ \ \ Assign the first and the second member of the memeplex as $\beta$ and $\delta$, respectively;
		\STATE \quad \quad \quad \quad \textbf{for} $j=1$ to $n$; // $n$ number members in each memeplex
		\STATE \ \ \ \ \ \ \ \ \textbf{Repeat} (for each member in current memeplex)
		\STATE \ \ \ \ \ \ \ \ \ \ \ \ Randomly select a solution from archive as $\alpha$;
		\STATE \ \ \ \ \ \ \ \ \ \ \ \ Update the position of the member by using $\alpha$, $\beta$ and $\delta$;
		\STATE \ \ \ \ \ \ \ \ \ \ \ \ Apply the crossover operation;
		\STATE \ \ \ \ \ \ \ \ \ \ \ \ Apply Levy flight protocol;
		\STATE \ \ \ \ Gather and shuffle the memeplexes as new population;
		\STATE \ \ \ \ Separate the population into frontiers;
		\STATE \ \ \ \ Update the archive according to the new population;
		\STATE \textbf{Return} the archive;
	\end{algorithmic}
	\label{algo:03}
\end{algorithm}

\section{Experiments}
In this section, we first assess MSGW-FLM. Next, we design two material allocation examples with multiple objectives and employ MSGW-FLM to solve the associated decision-making models. Notably, for dynamic decision scenarios, we apply a rolling time-domain planning approach \cite{roolcuisinier2022new}. This splits the entire decision timeline into multiple cycles, each tackled by MSGW-FLM, with the resulting plan being executed. When new data emerges, it's incorporated into the next cycle, allowing dynamic adjustments to model parameters.

The rolling planning window is set at 1 day for planning and 2 days for the planning domain. As for MSGW-FLM's parameters, we configure a population size $P=100$, maximum archive size $ArcMax=100$, max fitness evaluation count $MaxFit=20000$, memeplex count $m=5$, crossover rate $CR=0.4$, and random flight strategy parameter $\beta=1.4$.

\begin{table*}[!t]
\tiny
\caption{HV, IGD, and Spread performance of the optimized algorithm on ZDT, WFG, DTLZ, and LZ09\_F.} \label{tab:01}
\centering
\begin{tabular}{c|c|cccc|cccc|cccc}
\hline
\multirow{2}{*}{\textbf{No.}} & \multirow{2}{*}{\textbf{Problem}} & \multicolumn{4}{c|}{\textbf{HV}} & \multicolumn{4}{c|}{\textbf{IGD}} & \multicolumn{4}{c}{\textbf{Spread}} \\ \cline{3-14} 
&&\textbf{NSGA-II} & \textbf{IBEA} &  \textbf{MOEA/D} & \textbf{MSGW-FLM}    &\textbf{NSGA-II} & \textbf{IBEA} &  \textbf{MOEA/D} & \textbf{MSGW-FLM}     & \textbf{NSGA-II} & \textbf{IBEA} &  \textbf{MOEA/D} & \textbf{MSGW-FLM}      \\ \hline

01 & ZDT1 & 0.659 & \cellcolor{cyan}\textbf{0.661} & 0.641 & \cellcolor{cyan}\textbf{0.661} & 1.87e-4 & 1.64e-4 & 5.31e-4 & \cellcolor{cyan}\textbf{1.34e-4}  & 0.368 & 0.297 & 0.366 & \cellcolor{cyan}\textbf{0.0798} \\
02 & ZDT2 & 0.326 & 0.327 & 0.311 & \cellcolor{cyan}\textbf{0.328} & 1.94e-4 & 5.42e-4 & 4.40e-4 & \cellcolor{cyan}\textbf{1.41e-4} & 0.381 & 0.336 & 0.325 & \cellcolor{cyan}\textbf{0.0751} \\
03 & ZDT3 & 0.515 & 0.509 & 0.442 & \cellcolor{cyan}\textbf{0.516} & 1.33e-4 & 1.51e-3 & 1.7e-3 & \cellcolor{cyan}\textbf{1.01e-4} & 0.748 & 0.118 & 0.994 & \cellcolor{cyan}\textbf{0.706} \\
04 & ZDT4 & 0.655 & 0.236 & 0.301 & \cellcolor{cyan}\textbf{0.661} & 2.84e-4 & 2.23e-2 & 1.10e-2 & \cellcolor{cyan}\textbf{1.35e-4} & 0.388 & 1.112 & 0.969 & \cellcolor{cyan}\textbf{0.0878} \\
05 & ZDT6 & 0.389 & 0.396 & \cellcolor{cyan}\textbf{0.401} & \cellcolor{cyan}\textbf{0.401} & 3.27e-4 & 2.55e-4 & 1.42e-4 & \cellcolor{cyan}\textbf{1.33e-4} & 0.364 & 0.415 & 0.154 & \cellcolor{cyan}\textbf{0.0655} \\
\hline

06 & WFG1 & \cellcolor{cyan}\textbf{0.519} & 0.472 & 0.322 & 0.140 & \cellcolor{cyan}\textbf{3.16e-3} & 3.67e-3 & 6.70e-3 & 1.21e-2  & \cellcolor{cyan}\textbf{0.721} & 0.873 & 1.071 & 0.886 \\
07 & WFG2 & \cellcolor{cyan}\textbf{0.554 }& 0.549 & \cellcolor{cyan}\textbf{0.554} & 0.553 & 1.98e-3 & 3.76e-3 & 4.38e-4 & \cellcolor{cyan}\textbf{3.45e-4}  & \cellcolor{cyan}\textbf{0.791} & 1.24 & 1.112 & 0.958 \\
08 & WFG3 & 0.492 & \cellcolor{cyan}\textbf{0.494} & 0.493 & 0.491 & 1.55e-4 & \cellcolor{cyan}\textbf{1.30e-4} & 1.36e-4 & 1.36e-4  & 0.374 & 0.255 & 0.345 & \cellcolor{cyan}\textbf{0.119} \\
09 & WFG4 & 0.208 & \cellcolor{cyan}\textbf{0.209} & 0.204 & \cellcolor{cyan}\textbf{0.209} & \cellcolor{cyan}\textbf{1.65e-4} & 5.22e-4 & 2.28e-4 & 1.78e-4  & 0.376 & 0.513 & 0.510 & \cellcolor{cyan}\textbf{0.239} \\
10 & WFG5 & 0.218 & 0.216 & 0.218 & \cellcolor{cyan}\textbf{0.221}  & 1.29e-4 & 3.71e-4 & 1.24e-4 & \cellcolor{cyan}\textbf{1.05e-4}  & 0.410 & 0.585 & 0.453 & \cellcolor{cyan}\textbf{0.136} \\
11 & WFG6 & 0.199 & 0.197 & \cellcolor{cyan}\textbf{0.209} & 0.206  & 3.91e-4 & 8.54e-4 & \cellcolor{cyan}\textbf{2.28e-4} & 2.33e-4  & 0.387 & 0.526 & 0.413 & \cellcolor{cyan}\textbf{0.224} \\
12 & WFG7 & \cellcolor{cyan}\textbf{0.210} & 0.207 & 0.208 & 0.207  & 1.57e-4 & 5.14e-4 & \cellcolor{cyan}\textbf{1.48e-4} & 1.62e-4  & 0.349 & 0.515 & 0.414 & \cellcolor{cyan}\textbf{0.242} \\
13 & WFG9 & 0.224 & 0.224 & 0.221 & \cellcolor{cyan}\textbf{0.225}  & 2.34e-4 & 4.83e-4 & 2.41e-4 & \cellcolor{cyan}\textbf{1.87e-4}  & 0.367 & 0.501 & 0.448 & \cellcolor{cyan}\textbf{0.159} \\
\hline

14 & DTLZ1 & \cellcolor{cyan}\textbf{0.622} & 0.174 & 0.201 & 0.0567  & \cellcolor{cyan}\textbf{1.76e-3} & 4.17e-3 & 7.94e-3 & 1.21e-2  & 0.909 & 1.631 & 1.091 & \cellcolor{cyan}\textbf{0.681} \\
15 & DTLZ2 & 0.375 & \cellcolor{cyan}\textbf{0.412} & 0.0755 & 0.382 & 7.71e-4 & 1.40e-3 & 5.72e-3 & \cellcolor{cyan}\textbf{6.96e-4}  & 0.694 & \cellcolor{cyan}\textbf{0.582} & 1.001 & 0.627 \\
16 & DTLZ4 & 0.374 & 0.243 & 0.0814 & \cellcolor{cyan}\textbf{0.383} & 1.22e-3 & 5.11e-3 & 1.16e-2 & \cellcolor{cyan}\textbf{1.13e-3}  & 0.674 & 0.704 & 1.021 & \cellcolor{cyan}\textbf{0.641} \\
17 & DTLZ5 & 0.0928 & 0.0916 & 0.0143 & \cellcolor{cyan}\textbf{0.0933} & \cellcolor{cyan}\textbf{2.03e-5} & 1.02e-4 & 1.48e-3 & 2.74e-5  & 0.448 & 0.674 & 1.001 & \cellcolor{cyan}\textbf{0.304} \\
18 & DTLZ6 & 0.0001 & 0.0722 & 0.0145 & \cellcolor{cyan}\textbf{0.0951} & 7.15e-3 & 4.49e-4 & 3.82e-3 & \cellcolor{cyan}\textbf{3.39e-5} & 0.815 & 0.983 & 1.000 & \cellcolor{cyan}\textbf{0.103} \\
19 & DTLZ7 & 0.278 & 0.234 & 0.0638 & \cellcolor{cyan}\textbf{0.289} & 2.28e-3 & 1.59e-2 & 2.84e-2 & \cellcolor{cyan}\textbf{2.93e-3} & 0.756 & 0.826 & 0.991 & \cellcolor{cyan}\textbf{0.673} \\
\hline
20 & LZ09\_F1 & 0.652 & 0.654 & \cellcolor{cyan}\textbf{0.661} & 0.638 & 4.34e-4 & 6.84e-4 & \cellcolor{cyan}\textbf{2.36e-4} & 8.95e-4  & 0.508 & 0.766 & \cellcolor{cyan}\textbf{0.313} & 0.646 \\
21 & LZ09\_F2 & 0.507 & 0.492 & \cellcolor{cyan}\textbf{0.527} & 0.519 & 6.38e-3 & 8.26e-3 & \cellcolor{cyan}\textbf{4.89e-3} & 5.96e-3  & 1.460 & 1.471 & \cellcolor{cyan}\textbf{0.991} & 1.41 \\
22 & LZ09\_F3 & \cellcolor{cyan}\textbf{0.596} & 0.584 & \cellcolor{cyan}\textbf{0.596} & 0.564 & \cellcolor{cyan}\textbf{3.22e-3} & 6.65e-3 & 5.16e-3 & 4.09e-3  & 0.728 & 1.121 & \cellcolor{cyan}\textbf{0.701} & 0.981 \\
23 & LZ09\_F4 & 0.609 & 0.601 & \cellcolor{cyan}\textbf{0.621} & 0.574 & 3.19e-3 & 6.10e-3 & \cellcolor{cyan}\textbf{2.98e-3} & 4.38e-3 & \cellcolor{cyan}\textbf{0.577} & 1.031 & 0.971 & 0.757 \\
24 & LZ09\_F5 & \cellcolor{cyan}\textbf{0.609} & 0.602 & \cellcolor{cyan}\textbf{0.609} & 0.583 & \cellcolor{cyan}\textbf{2.74e-3} & 4.58e-3 & 4.03e-3 & 3.49e-3  & \cellcolor{cyan}\textbf{0.641} & 1.092 & 0.665 & 0.873 \\
25 & LZ09\_F6 & 0.163 & 0.0642 & 0.0725 & \cellcolor{cyan}\textbf{0.208} & \cellcolor{cyan}\textbf{7.32e-3} & 1.62e-2 & 1.92e-2 & 7.35e-3  & 0.937 & 1.671 & 0.987 & \cellcolor{cyan}\textbf{0.844} \\
26 & LZ09\_F7 & \cellcolor{cyan}\textbf{0.447} & 0.393 & 0.443 & 0.206 & 1.09e-2 & 1.92e-2 & \cellcolor{cyan}\textbf{9.18e-3} & 1.93e-2  & 1.38 & 1.121 & 1.181 & \cellcolor{cyan}\textbf{0.936} \\
27 & LZ09\_F8 & \cellcolor{cyan}\textbf{0.406} & 0.385 & 0.349 & 0.181 & \cellcolor{cyan}\textbf{1.03e-2} & 1.69e-2 & 1.06e-2 & 1.76e-2  & 1.25 & 1.118 & 1.271 & \cellcolor{cyan}\textbf{0.799} \\
28 & LZ09\_F9 & 0.164 & 0.162 & 0.166 & \cellcolor{cyan}\textbf{0.193} & 8.19e-3 & 9.02e-3 & \cellcolor{cyan}\textbf{4.80e-3} & 6.15e-3  & 1.58 & 1.672 & \cellcolor{cyan}\textbf{0.976} & 1.571  \\
\hline
\multicolumn{2}{c|}{\textbf{No. of Optimal Indicator}} & 8 & 4 & 8 & \cellcolor{cyan}\textbf{14} & 8 & 1 & 7 & \cellcolor{cyan}\textbf{12} & 4 & 1 & 4 & \cellcolor{cyan}\textbf{19}\\
\hline 
\end{tabular}
\end{table*}

\begin{figure*}
    \centering
    \begin{subfigure}{0.31\textwidth}
        \centering
        \includegraphics[width=\linewidth]{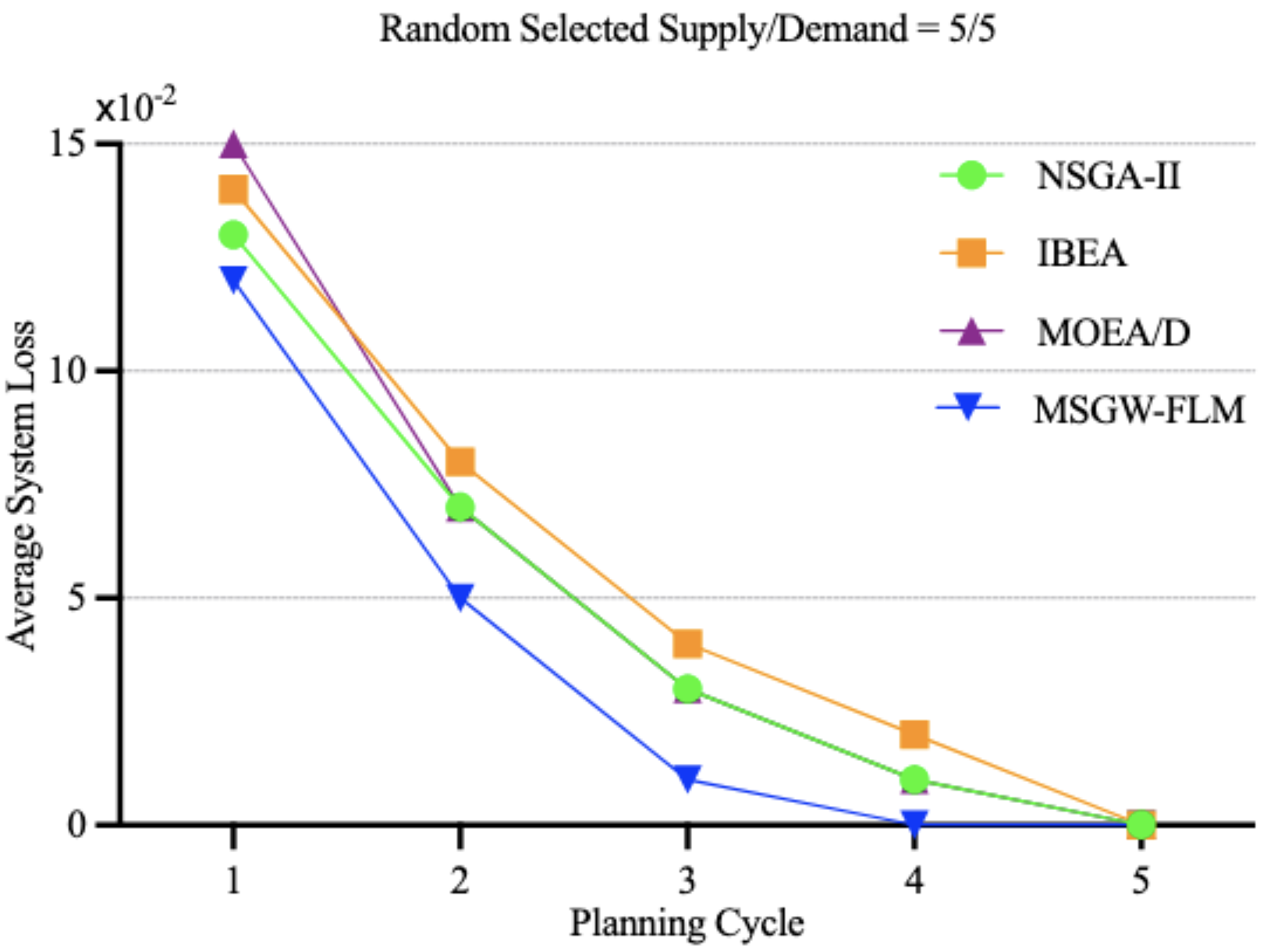}
        \caption{5/5}
        \label{fig:06}
    \end{subfigure}
    \begin{subfigure}{0.31\textwidth}
        \centering
        \includegraphics[width=\linewidth]{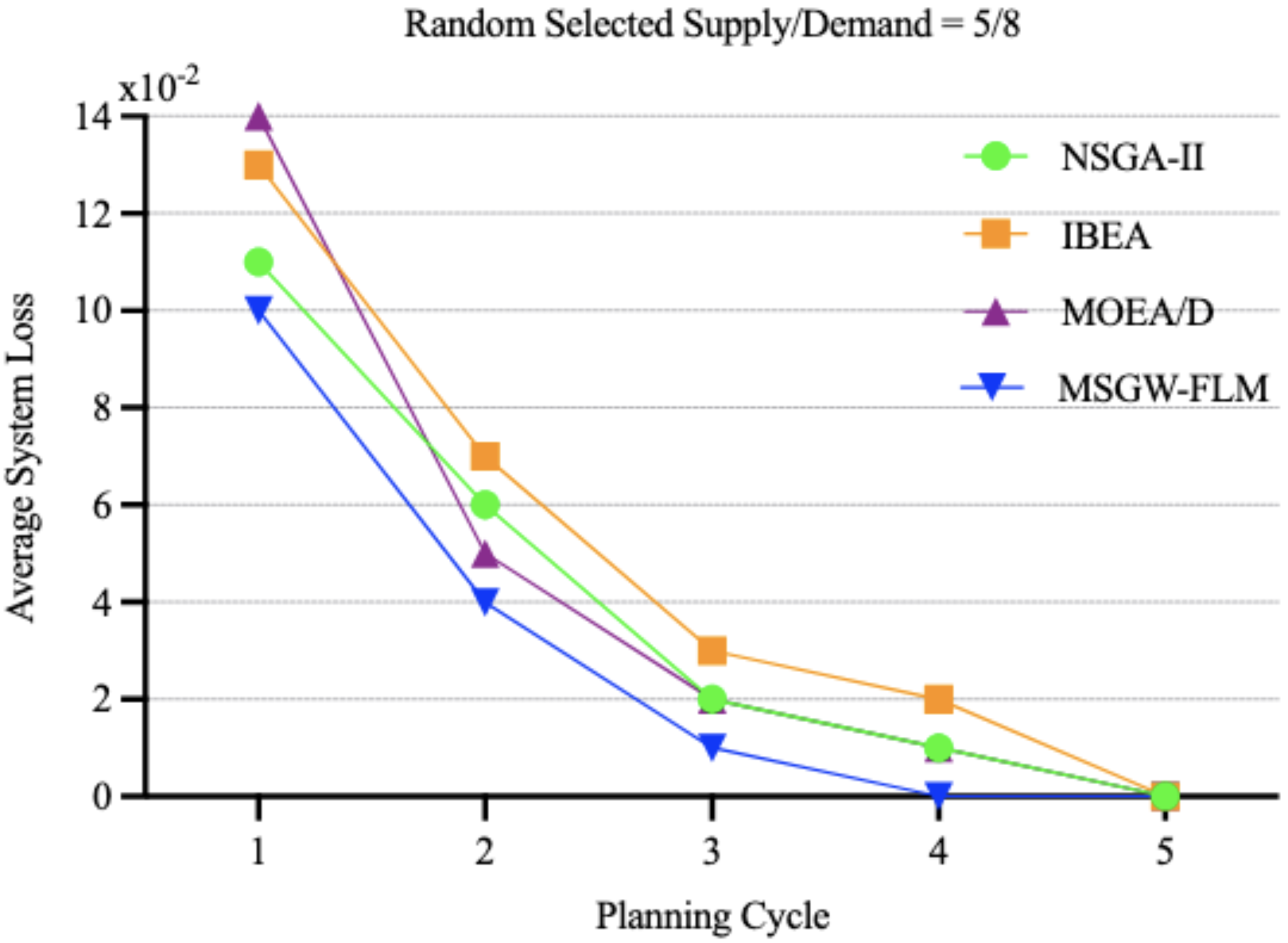}
        \caption{5/8}
        \label{fig:07}
    \end{subfigure}
    \begin{subfigure}{0.31\textwidth}
        \centering
        \includegraphics[width=\linewidth]{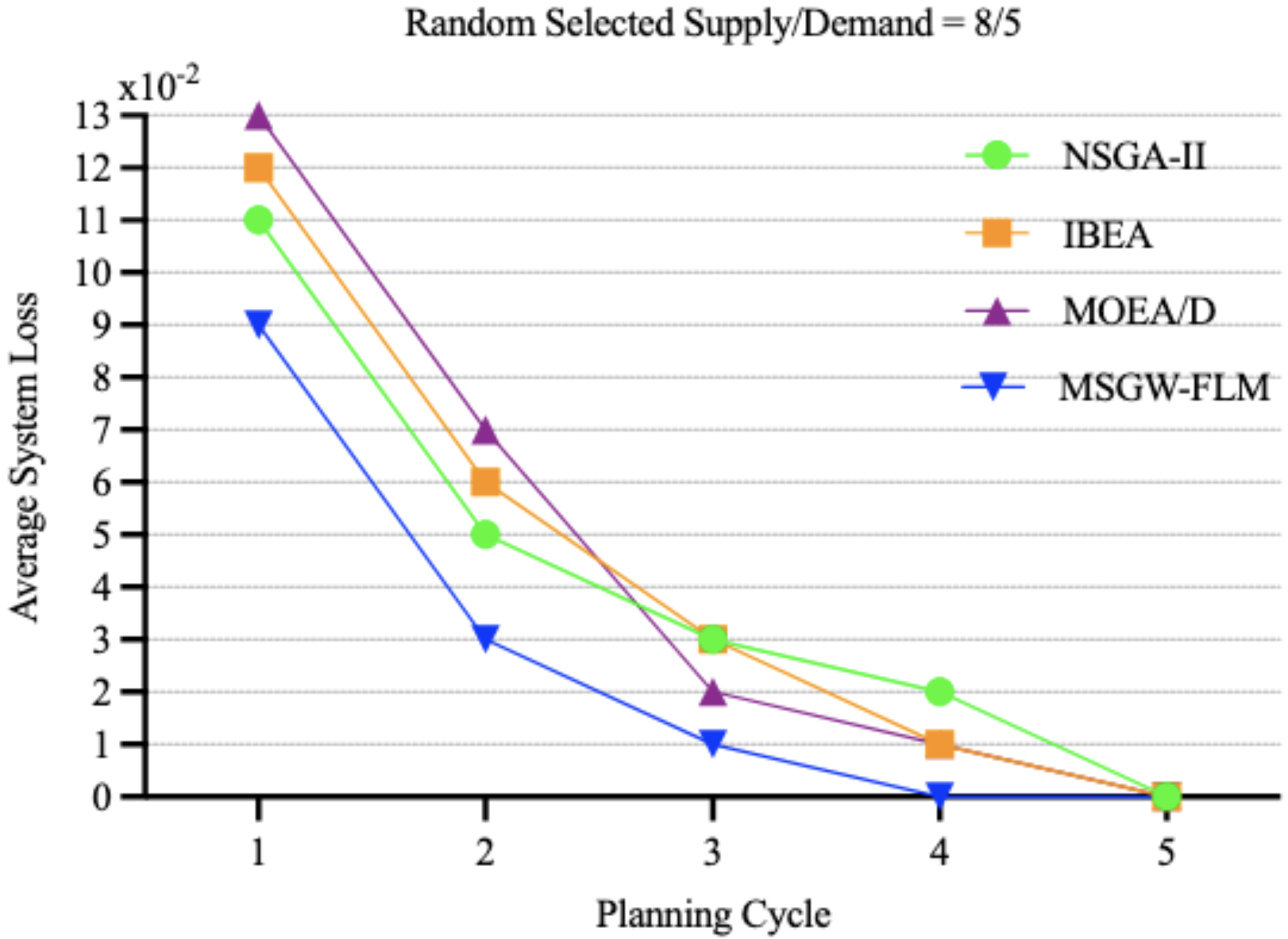}
        \caption{8/5}
        \label{fig:08}
    \end{subfigure}
    \caption{Average Loss for Disaster Sites per Period of Randomly Selected Supply/Demand Points.}
    \label{fig:main}
\end{figure*}

\subsection{MSGW-FLM Multi-objective Optimization Algorithm}

In this study, we compare the proposed MSGW-FLM with NSGA-II \cite{deb2002fast}, IBEA \cite{zitzler2004indicator}, and MOEA/D \cite{zhang2007moea}. We evaluate using four classic benchmark problems (ZDT, WFG, DTLZ, LZ09\_F) \cite{zhang2007moea,ozkics2017novel}, totaling 28 multi-objective sub-problems. 
Each algorithm was run 100 times, and the average value was used as the final assessment metric displayed in table \ref{tab:01}.
Performance metrics used are HV \cite{zitzler1999multiobjective}, IGD \cite{durillo2011jmetal}, and Spread \cite{babalik2018multi,xiang2015elitism}. For HV, larger values are better, while for IGD and Spread, smaller values are preferred.
Indicators with optimal performance are accentuated in cyan.
MSGW-FLM outperforms in half of the benchmark tests using the HV metric.
IGD measures both convergence and diversity, and MSGW-FLM leads in nearly half of the problems.
MSGW-FLM has the best scores in over half of Spread, highlighting its success in applying the memeplex strategy and superior Spread performance.
Overall, on 28 multi-objective benchmark tests, MSGW-FLM proves effective in addressing multi-objective issues, often outperforming other baseline algorithms.

\subsection{MSGW-FLM Based on Random Supply-Demand Points}

To further assess MSGW-FLM's capability in managing emergency material allocation, we delve into scenarios with unpredictable demand-supply dynamics. We evaluate the model using random supply-demand points, specifically between distribution centers and disaster-stricken areas. In our setup, there are 10 designated disaster areas and 10 distribution centers across 5 planning cycles. Throughout each cycle, we dictate the ambiguous demand at every disaster site per planning phase and the uncertain supply of each distribution center. By using these randomly determined disaster locations and distribution centers, we can feed our models with these supply-demand points to address the allocation challenge.

The mean change curves of total system losses for various supply/demand point combinations are displayed in Fig. \ref{fig:main}. These represent scenarios with 5/5, 5/8, and 8/5 supply/demand points respectively. From the visual results, we observe that all algorithms - NSGA-II, IBEA, MOEA/D, and MSGW-FLM - show a trend of decreasing system average loss for material distribution, ultimately reaching zero. This demonstrates the system's ability to efficiently allocate emergency relief materials over time, progressively reducing the deficits faced by disaster-affected areas. Most notably, MSGW-FLM consistently outperforms the others, achieving a mean system loss of zero by the fourth planning cycle. These findings underscore MSGW-FLM's efficacy and suitability for managing emergency material allocation tasks.

\section{Conclusion}

This study investigates the use of spatio-temporal data in IoT for multi-cycle emergency resource allocation, taking into account multiple objectives and constraints. It introduces the Multi-Objective Shuffled Gray-Wolf Frog Leaping Model (MSGW-FLM), a pioneering approach for dynamic multi-cycle emergency response. Using spatio-temporal data, the model is tested on 28 multi-objective tasks. The MSGW-FLM consistently outperforms baseline models such as NSGA-II, IBEA and MOEA/D in many scenarios. Future work will validate the model's effectiveness in real-world emergency scenarios and test its ability to dynamically update spatio-temporal information.

\bibliographystyle{IEEEtran}
\bibliography{mybib}

\end{document}